%% file: paper.tex
\documentclass{article} 
\usepackage{arxiv_submission,times}
\usepackage{hyperref}
\usepackage{url}
\usepackage{graphicx}
\usepackage[title, titletoc]{appendix}
\usepackage{multirow}
\usepackage{hyperref}

\title{Who wrote this book? A challenge for e-commerce}

\author{B\'eranger Dumont, Simona Maggio, Ghiles Sidi Said \& Quoc-Tien Au \\
Rakuten Institute of Technology Paris \\
\texttt{\{beranger.dumont,simona.maggio\}@rakuten.com},\\
\texttt{\{ts-ghiles.sidisaid,quoctien.au\}@rakuten.com} \\
}

\begin{document}

\maketitle
%

%
\begin{figure}[!b]
  \includegraphics[width=\textwidth]{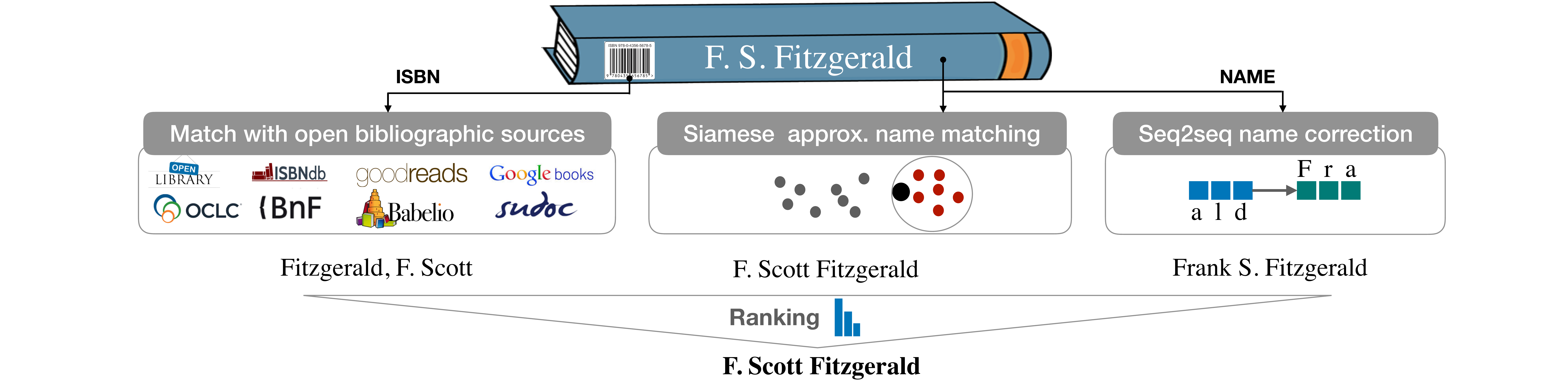}
  \caption{Overview of the system for normalizing author names. Each component is detailed in Section~\ref{expsetup}.}
  \label{fig:overview}
\end{figure}

%
\begin{abstract}
\input{abstract}
\end{abstract}

%

\section{Introduction}
\label{introduction}
\input{introduction}

\section{Book data}
\label{data}
\input{data}

\section{Experimental setup}
\label{expsetup}
\input{expsetup}

\section{Results}
\label{results}
\input{results}

\section{Related works}
\label{related}
\input{related}

\section{Conclusions}
\label{conclusions}
\input{conclusions}

\subsubsection*{Acknowledgments}
We thank Rapha\"el Ligier-Tirilly for his help with the deployment of the system as microservices, and Laurent Ach for his support.

%
\bibliographystyle{ACM-Reference-Format}
\bibliography{paper}

\end{document}

%% file: abstract.tex
Modern e-commerce catalogs contain millions of references, associated with textual and visual information that is of paramount importance for the products to be found via search or browsing.
Of particular significance is the book category, where the author name(s) field poses a significant challenge.
Indeed, books written by a given author (such as F.\ Scott Fitzgerald) might be listed with different authors' names in a catalog due to abbreviations and spelling variants and mistakes, among others.
To solve this problem at scale, we design a composite system involving open data sources for books as well as machine learning components leveraging deep learning-based techniques for natural language processing.
In particular, we use Siamese neural networks for an approximate match with known author names, and direct correction of the provided author's name using sequence-to-sequence learning with neural networks.
We evaluate this approach on product data from the e-commerce website Rakuten France, and find that the top proposal of the system is the normalized author name with 72\% accuracy.

%% file: introduction.tex
Unlike brick-and-mortar stores, e-commerce websites can list hundreds of millions of products, with thousands of new products entering their catalogs every day.
The availability and the reliability of the information on the products, or {\it product data}, is crucial for the products to be found by the users via textual or visual search, or using faceted navigation.

Books constitute a prominent part of many large e-commerce catalogs. Indeed, for the United States alone, more than 300,000 books are published every year and the market value of the book business is estimated to 274 billion euros in 2013~(\cite{ipa}).
Relevant book properties include: title, author(s), format, edition, and publication date, among others.
In this work, we focus on the names of book authors, as they are found to be extremely relevant to the user and are commonly used in search queries on e-commerce websites, but suffer from considerable variability and noise.
To the best of our knowledge, there is no large-scale public dataset for books that captures the variability arising on e-commerce marketplaces from user-generated input.
Thus, in this work we use product data from Rakuten France (RFR).\footnote{\url{https://fr.shopping.rakuten.com} \\ The dataset will be publicly available at \url{https://rit.rakuten.co.jp/data_release}.}

The variability and noise is evident in the RFR dataset. For example, books written by F.\ Scott Fitzgerald are also listed with the following author's names: ``Francis Scott Fitzgerald'' (full name), ``Fitzgerald, F.\ Scott'' (inversion of the first and last name), ``Fitzgerald'' (last name only), ``F.\ Scott Fitgerald'' (misspelling of the last name), ``F SCOTT FITZGERALD'' (capitalization and different typological conventions), as well as several combinations of those variations.
Less frequently, the author name field is missing (displayed as ``Unknown'') or incorrectly refers to a different person.

The variability of the possible spellings for an author's name is very hard to capture using rules, even more so for names which are not primarily written in latin alphabet (such as arabic or asian names), for names containing titles (such as ``Dr.'' or ``Pr.''), and for pen names which may not follow the usual conventions.
This motivated us to explore automated techniques for normalizing the authors' names to their best known (``canonical'') spellings.
In addition to improvements to the product data, normalizing the authors' names can be used to help the user find other books by the same author.

Fortunately, a wealth of open databases exist for books, making it possible to match a significant fraction of the books listed in e-commerce catalogs.
While not always clean and unambiguous, this information is extremely valuable in itself.
Furthermore, along with datasets of name variants, it enables us to train and evaluate machine learning systems to disambiguate and normalize authors' names.
To this end, in addition to the match with open databases, we will explore two different approaches: approximate match with known authors' names using Siamese neural networks, and direct correction of the provided author's name using sequence-to-sequence learning with neural networks.
Then, we will present our machine learning approach to ranking the results.

The rest of the paper is organized as follows: we present the data from RFR and from the open databases in Section~\ref{data}, before turning to the experimental setup for the overall system and for each of its components in Section~\ref{expsetup}.
Finally, we give results in Section~\ref{results}, we present related works in Section~\ref{related}, and conclude in Section~\ref{conclusions}.

%% file: data.tex
\subsection{Rakuten France data}

The RFR dataset contains 12 million book references. The most relevant product data to help normalizing authors' names is:
\begin{itemize}
\item {\bf ISBN}\footnote{International Standard Book Number, see \url{https://www.isbn-international.org}} in 10 digit or 13 digit format;
\item {\bf product title}, which almost always includes the book title, often supplemented with extra information in free text;
\item {\bf author(s)} of the book as was provided by the seller (multiple authors should be, but not always are, listed in separate sub-fields).
\end{itemize}

In particular, the ISBN is a worldwide unique identifier for books, which makes it a prime candidate for unambiguous matching with external sources.
In this dataset, an ISBN is present for about 70\% of the books. Among the books with no ISBN, 30\% are ancient books which are not expected to be associated an ISBN.

\subsection{External bibliographic resources}
\label{opendata}

There is no central authority providing consistent information on books associated with an ISBN.
However, there is a wealth of bibliographic resources and open databases for books.
In order to retrieve the author's name(s) associated with the books in the RFR dataset, we perform ISBN matching using public APIs on eight of them, listed in Table~\ref{tab:isbn2} along with the the fraction of found ISBNs from this dataset.
%
The last three sources are specialized on French books and books translated into French, which is relevant for the RFR dataset where such books are overrepresented.
We find that 75\% of the books with an ISBN are matched with at least one source.
We also find the sources to be highly complementary in terms of coverage, and to be independent to a reasonable extent (i.e., returned results can differ significantly in case of match with different sources). 
The match via ISBN on external bibliographic resources is the first component of the system depicted in Fig.~\ref{fig:overview}.


\begin{table}[t]
  \caption{Performances of the external bibliographic resources used for matching books on RFR via ISBN.}
  \label{tab:isbn2}
  \begin{center}
  \begin{tabular}{lll}
    \hline
    Source & URL & \% of ISBNs \\
    \hline
    Open Library & \url{https://openlibrary.org} & 24.9\% \\
    ISBNdb & \url{https://isbndb.com} & 36.3\% \\
    Goodreads & \url{https://www.goodreads.com} & 64.7\% \\
    Google Books & \url{https://books.google.com} & 51.2\% \\
    OCLC & \url{https://www.oclc.org} & 52.2\% \\
    BnF & \url{http://www.bnf.fr} & 7.4\% \\
    Sudoc & \url{http://www.sudoc.abes.fr} & 29.0\% \\
    Babelio & \url{https://www.babelio.com} & 7.9\% \\
  \hline
\end{tabular}
\end{center}
\end{table}

\subsection{Dataset of name entities}
\label{data-entities}

In order to train and evaluate machine learning systems to match or correct authors' names, a dataset of name entities containing the different surface forms (or variants) of authors' names is required.
The entities should reflect as well as possible the variability that can be found in the RFR dataset, as was illustrated in the case of F.\ Scott Fitzgerald in Section~\ref{introduction}.
For each entity, a canonical name should be elected and correspond to the name that should be preferred for the purpose of e-commerce (i.e., its most popular variant). 

Name entities are collected in three distinct ways:
\begin{enumerate}
\item \textbf{ISBN matching}: books matched with external sources via ISBN may reveal author names different from one another, and/or compared with the RFR author name field, which gives us an entity. The canonical form is the one that is matched with Wikipedia\footnote{\url{https://www.wikipedia.org}} or DBpedia~(\cite{dbpedia}); else the one provided by the greatest number of sources.
\item \textbf{Matching of Rakuten authors}: we build entities using fuzzy search on the author name field on DBpedia and consider the DBpedia value to be canonical.
We limit the number of false positives in fuzzy search by tokenizing both names, and keeping only the names where at least one token from the name on RFR is approximately found in the external resource (i.e., with Levenshtein distance $< 2$).
\item \textbf{Name variants}: DBpedia, BnF, and JRC-names~(\cite{jrc1,jrc2}) directly provide data about people (not limited to book authors) and their name variants.
\end{enumerate}

While Wikipedia seems more pertinent to select canonical names matching the e-commerce user expectations, specialized librarian data services, such as the Library of Congress Name Authority\footnote{\url{http://id.loc.gov/authorities/names.html}}, could be used in future research to enrich the dataset of name entities.

After creating the name entity dataset, we normalize all names to latin-1.
We obtain about 750,000 entities, for a total of 2.1 million names.

\subsection{Annotated Rakuten France data}
\label{data-ranking}

In order to evaluate the overall system, we need product data from RFR for which the canonical author name has been carefully annotated and can be considered as the ground truth.
To this end, we have considered a subset of 1000 books from the RFR dataset, discarding books written by more than one author for simplicity.\footnote{The two RFR datasets (original 12 million books and 1000 annotated books) will be publicly available at \url{https://rit.rakuten.co.jp/data_release}.}
We find that 467 books have a canonical author name that differs from RFR's original (unnormalized) author name.
Also, 310 do not have an ISBN or do not match on any of the bibliographic resources listed in Section~\ref{opendata}.
Among them, 208 books have a canonical name that differs from the catalog input.

%% file: expsetup.tex
The overview of the system can be found in Fig.~\ref{fig:overview}.
Its first component, the matching via ISBN against external databases, has already been presented in Section~\ref{opendata}.
In the rest of this section, we will shed light on the three machine learning components of the system.

\subsection{Siamese approximate name matching}

We want to learn a mapping that assigns a similarity score to a pair of author names such that name variants of the same entity will have high similarity, and names that belong to different entities will have low similarity.
To this end, we might use a classical string metric such as the Levenshtein distance or the $n$-gram distance~(\cite{kondrak2005n}).
However, those are not specific to people's names, and might return a large distance (low similarity) in cases such as the inversion between first name and last name or the abbreviation of the first name to an initial.
Thus, we want to use the dataset of name entities to learn a specialized notion of similarity---this is known as distance metric learning~(\cite{kulis2013metric}).
Once learned, this mapping will enable us to assign an entity to any given name.

To learn such a mapping, we use a pair of neural networks with shared weights, or Siamese neural network~(\cite{siamese}).
Each network is a recurrent neural network (RNN) composed of a character-level embedding layer with 256 units, a bidirectional long short-term memory (LSTM)~(\cite{lstm}) with $2 \times 128$ units, and a dense layer with 256 units.
Each network takes a name as input and outputs a representation---the two representations are then compared using cosine similarity.

The input to the Siamese network consists of a tuple $(x_1, x_2, y)$, where $x_i$ are the input names viewed as a sequence of characters, and the similarity $y$ is equal to 1 for name variants of the same entity, and to 0 otherwise.
We preprocess the input by representing all characters in ASCII and lowercase. We consider a sequence length of 32 using zero padding.

The weights of the embedding, of the LSTM, and of the dense layers are learned during training.
The Siamese network is trained with contrastive loss~(\cite{contrastive}) in order to push the similarity towards 1 for similar pairs, and below a certain margin (that we set to 0) for dissimilar pairs.
The optimization is done using Adam~(\cite{adam}), with a learning rate of $10^{-3}$ and a gradient clipping value of 5.
We use batches of 512 samples, consider a negative to positive pairs ratio of $4:1$, and randomly generate new negative pairs at every epoch.
The architecture is implemented, trained, and tested using Keras\footnote{https://github.com/keras-team/keras}.

At test time, we search for the canonical name whose representation is closest to that of the query, using only the high-quality name entities from DBpedia, BnF, and JRC-names. To this end, we do approximate nearest neighbor search using Annoy\footnote{https://github.com/spotify/annoy}.

\subsection{Name correction with seq2seq networks}

We use a generative model to correct and normalize authors' names directly. The dataset of name entities is again employed to train a sequence-to-sequence (seq2seq) model~(\cite{seq2seq}) using Keras to produce the canonical form of a name from one of its variants. The dataset is further augmented by including additional variants where the first name is abbreviated to an initial.

The seq2seq model is an encoder-decoder using RNNs, with a character embedding layer, as in the case of the Siamese network. The encoder is a bi-directional LSTM with $2 \times 256$ units, while the decoder is a plain LSTM with $512$ units connected to a softmax layer that computes a probability distribution over the characters.

The training is performed by minimizing the categorical cross-entropy loss, using teacher forcing~(\cite{teacher}) for faster optimization. The optimization setting is identical to that of the Siamese nework, with batches of $1024$ samples.
For inference, we collect the $10$ output sequences with highest probability using beam search.

\subsection{Ranking of the proposals}

For any given book with an ISBN and an author's name, all three techniques shown in Fig.~\ref{fig:overview} provide one or several candidate canonical names.
As we aim at providing an automated tool to enhance the quality of the book products, the final system should provide a ranked list of candidates with a calibrated confidence level.
For this purpose, we evaluate each candidate canonical name by using logistic regression on a set of 12 features related to their origin.

More specifically, we have binary features for each candidate: $8$ indicating whether it is found in the bibliographic sources, one indicating equality with the input name, one indicating if found in the top-10 candidates of the seq2seq model, and one indicating if found by approximate match with the Siamese network. We also have a feature corresponding to the cosine distance between the representation of the proposal and that of the input name.
For a given proposal, the selected features reflect that the confidence of the global system should increase with {\it (i)} the consensus among the different sources, and {\it (ii)} the similarity to the input catalog name.

We train and evaluate a logistic regression on the annotated dataset introduced in Section~\ref{data-ranking} to estimate the probability that a proposal is the canonical form for an author's name.
This information can then be used to rank all the generated canonical names for a given book.
For each book, about $10$ proposals are generated, making a dataset of 11185 samples.
We split the books between training and test sets, with a ratio of $50\% : 50\%$. As the negative proposals greatly outnumber the positive ones, we use random oversampling on the training data.


%% file: results.tex
The three machine learning components discussed in the previous section have been individually evaluated on their specific task. Furthermore the final system has been evaluated in terms of correctly normalized book authors in a real case scenario. 

\subsection{Siamese approximate name matching}

We evaluate the Siamese network on a held out test set, and compare it to an $n$-gram distance, by checking that the nearest neighbor of a name variant is the canonical name of the entity to which it belongs.
We find an accuracy of $79.8\%$ for the Siamese network, against $71.1\%$ for the $n$-gram baseline with $n=3$.
We have also checked metrics when introducing a threshold distance above which we consider that no matching entity is found, and found systematic improvement over the baseline.
In the final system, we set the threshold to infinity.

\subsection{Name correction with seq2seq networks}

Similarly to the previous approach, the seq2seq network is evaluated on a held out test set by checking that one of the generated name variants is the canonical name of the entity to which it belongs.
As expected, name normalization using seq2seq network gives poorer performances than approximate matching within a dataset of known authors, but constitutes a complementary approach that is useful in case of formatting issues or incomplete names. This approach alone reaches a top-$10$ accuracy of $42\%$ on the entire test set, $26\%$ on a test set containing only names with initials, and $53\%$ on a test set containing only minor spelling mistakes.

\subsection{Ranking of the proposals}

A logistic regression is trained to predict whether a proposed name variant for a given input name is the actual canonical author name, according to features related to the sources providing the proposal, as well as the similarity to the input name.
With a decision threshold of $p=0.5$, the trained classifier has an accuracy of 93\% for both positive and negative candidates in the test set.
The coefficients of the estimator reveal the importance of the features and, thus, of the related components. We observe that the three most important contributions are the match with the Siamese network, the match via ISBN in Babelio, and the similarity with the input name in the catalog, confirming the relevance of a multi-approach design choice.

Finally, the probability that a proposed name variant is the canonical author name, provided by the resulting classifier, is used as a confidence score to rank the different candidate names returned by the three normalization approaches.   

\subsection{Global system}

In order to reflect the actual use of the global system on e-commerce catalog data, the final evaluation is performed at the book level, by considering all the proposals provided by the different components for a given book.
The metric used is the top-$k$ accuracy on the ranked list of proposals for each book; results are summarized in Table~\ref{tab:bookres}.
We find that $72\%$ of the books have the author's name normalized by the highest ranked proposal.
Excluding from the evaluation books where the ground truth for the author's name equals the catalog value, this accuracy drops to $49\%$.
In the case of books without ISBN or that do not match on any of the bibliographic resources, thus relying on machine learning-based components only, we find that $50\%$ of the books are normalized by the top proposal.
Finally, for the combination of the above two restrictions, we find a top-1 accuracy of 35\%.
 
\begin{table}[t]
  \caption{Global system top-$k$ accuracy at the book level.}
  \label{tab:bookres}
  \begin{center}
  \begin{tabular}{llll}
    \hline
    Type of books & \# of samples & acc@1 & acc@3 \\
    \hline
     all & 500 & 72\% & 85\% \\
     unnormalized input author & 235 & 49\% & 67\% \\
     no ISBN match & 151 & 50\% & 64\% \\
     unnormalized + no ISBN & 109 & 35\% & 49\% \\
  \hline
  \vspace*{-10mm}
\end{tabular}
\end{center}
\end{table}

%% file: related.tex

There is a long line of work on author name disambiguation for the case of bibliographic citation records~(\cite{namereview2}).
While related, this problem differs from the one of book authors.
Indeed, unlike most books, research publications usually have several authors, each of them having published papers with other researchers.
The relationships among authors, which can be represented as a graph, may be used to help disambiguate the bibliographic citations.
In addition, unlike books in the context of e-commerce, the author name field is usually clean.

Named entity linking~(\cite{entlink2}), where one aims at determining the identity of entities (such as a person's name) mentioned in text, is another related problem.
This approach requires a knowledge base to which entity mentions can be linked.
The crucial difference with the disambiguation of book authors is that entity linking systems leverage the context of the named entity mention to link unambiguously to an entity in the knowledge base.

Deep learning approaches similar to ours have been used to tackle related problems.
Distance metric learning with neural networks has been used for merging datasets on names~(\cite{mergingdatasets}), for normalizing job titles~(\cite{jobtitles}), and for the disambiguation of researchers~(\cite{aminer}). 
Sequence-to-sequence learning has been used for the more general task of text normalization~(\cite{textnorm}), and for sentence-level grammar error identification~(\cite{schmaltz2016sentence}).

To the best of our knowledge, the problem of name normalization specific to book authors has not been tackled in the previous literature, with the exception of a work on named entity linking for French writers~(\cite{nel-frbooks}).

%% file: conclusions.tex
We provided a first attempt at solving the problem of author name normalization in the context of books sold on e-commerce websites.
To this end, we used a composite system involving open data sources for books, approximate match with Siamese neural networks, name correction with sequence-to-sequence networks, and ranking of the proposals.\footnote{The technical debt of such machine learning systems~(\cite{hiddentech}) should be kept in mind.}
We find that 72\% of the books have the author's name normalized by the highest ranked proposal.

In order to facilitate future research, we are releasing data from Rakuten France: a large dataset containing product information, and a subset of it with expert human annotation for the authors' names.
They will be accessible at \url{https://rit.rakuten.co.jp/data_release}.

Multiple challenges remain and are left for future research.
First, the system should be extended to handle the case of books with multiple authors.
In addition, 
the book title could be used to help disambiguate between authors and to query external bibliographic resources.
This work can also be seen as an intermediate step towards a knowledge base for book author names with name variants, extending public ones such as BnF and DBpedia, using the ISNI\footnote{International Standard Name Identifier, see \url{http://www.isni.org}} for easier record linkage whenever available.